\documentclass[11pt,a4paper]{article}
\pdfoutput=1
\usepackage[hyperref]{acl2020}
\usepackage{times}
\usepackage{latexsym}

\usepackage{microtype}

\usepackage{rotating}
\usepackage{blindtext}
\usepackage{amsmath}
\usepackage{amsfonts}
\usepackage{graphicx}
\usepackage{mathtools}
\usepackage{booktabs}
\usepackage{relsize}
\usepackage{float}

\usepackage{threeparttable}
\usepackage{booktabs, caption, makecell}

\usepackage[title]{appendix}
\usepackage[cal=cm]{mathalfa}
\usepackage[colorinlistoftodos]{todonotes}
\usepackage{makecell, multirow,}
\usepackage[export]{adjustbox}
\usepackage{url}
\usepackage{textcomp}

\usepackage{fancyhdr}
\newcommand{\mycomment}[1]{}                     
\newcommand{\ignore}[1]{}

\aclfinalcopy 
\def\aclpaperid{1501} 

\title{
  The Unstoppable Rise of Computational Linguistics in Deep Learning
}

\author{James Henderson\\
  Idiap Research Institute, Switzerland \\
  \texttt{james.henderson@idiap.ch} \\
}

\date{}

\begin{document}
\thispagestyle{fancy}%

\maketitle

\begin{abstract}
In this paper, we trace the history of neural networks applied to natural language understanding tasks, and identify key contributions which the nature of language has made to the development of neural network architectures.  We focus on the importance of variable binding and its instantiation in attention-based models, and argue that Transformer is not a sequence model but an induced-structure model.  This perspective leads to predictions of the challenges facing research in deep learning architectures for natural language understanding.
\end{abstract}

\section{Introduction}

When neural networks first started being applied to natural language in the 1980s and 90s, they represented a radical departure from standard practice in computational linguistics.  Connectionists had vector representations and learning algorithms, and they didn't see any need for anything else.  Everything was a point in a vector space, and everything about the nature of language could be learned from data.  On the other hand, most computational linguists had linguistic theories and the poverty-of-the-stimulus argument.  Obviously some things were learned from data, but all the interesting things about the nature of language had to be innate.

A quarter century later, we can say two things with certainty: they were both wrong.  Vector-space representations and machine learning algorithms are much more powerful than was thought.  Much of the linguistic knowledge which computational linguists assumed needed to be innate can in fact be learned from data.  But the unbounded discrete structured representations they used have not been replaced by vector-space representations.  Instead, the successful uses of neural networks in computational linguistics have replaced specific pieces of computational-linguistic models with new neural network architectures which bring together continuous vector spaces with structured representations in ways which are novel for both machine learning and computational linguistics.

Thus, the great progress which we have made through the application of neural networks to natural language processing should not be viewed as a conquest, but as a compromise.  As well as the unquestionable impact of machine learning research on NLP, the nature of language has had a profound impact on progress in machine learning.  In this paper we trace this impact, and speculate on future progress and its limits.

We start with a sketch of the insights from grammar formalisms about the nature of language, with their multiple levels, structured representations and rules.  The rules were soon learned with statistical methods, followed by the use of neural networks to replace symbols with induced vectors, but the most effective models still kept structured representations, such as syntactic trees.  More recently, attention-based models have replaced hand-coded structures with induced structures.  The resulting models represent language with multiple levels of structured representations, much as has always been done.  Given this perspective, we identify remaining challenges in learning language from data, and its possible limitations.

\section{Grammar Formalisms versus Connectionism} 
\label{sec:reps}

\subsection{Grammar Formalisms}

Our modern understanding of the computational properties of language started with the introduction of grammar formalisms.
Context Free Grammars \citep{Chomsky59} illustrated how a formal system could model the infinite generative capacity of language with a bounded grammar.  This formalism soon proved inadequate to account for the diversity of phenomena in human languages, and a number of linguistically-motivated grammar formalisms were proposed (e.g HPSG \citep{Pollard87}, TAG \citep{Joshi87}, CCG \citep{Steedman00}).

All these grammar formalisms shared certain properties, motivated by the understanding of the nature of languages in Linguistics.  They all postulate representations which decompose an utterances into a set of sub-parts, with labels of the parts and a structure of inter-dependence between them.  And they all assume
that this decomposition happens at multiple levels of representation.  For example that spoken utterances can be decomposed into sentences, sentences can be decomposed into words, words can be decomposed into morphemes, and morphemes can be decomposed into phonemes, before we reach the observable sound signal.  In the interests of uniformity, we will refer to the sub-parts in each level of representation as its {\em entities}, their labels as their {\em properties}, and their structure of inter-dependence as their {\em relations}.  The
structure of
inter-dependence between entities at different levels will also be referred to as relations.

In addition to these representations, grammar formalisms include specifications of the allowable structures.  These may take the form of hard constraints or soft objectives, 
or of deterministic rules or stochastic processes.
In all cases,
the purpose of these specifications is to account for the regularities found in natural languages.  
In the interests of uniformity, we will refer to all these
different
kinds of specifications of allowable structures as {\em rules}.  These rules may apply within or between levels of representation.

In addition to explicit rules, computational linguistic formalisms implicitly make claims about the regularities found in natural languages through their expressive power.  Certain types of rules simply cannot be specified, thus claiming that such rules are not necessary to capture the regularities found in any natural language.  These claims differ across formalisms, but the study of the expressive power of grammar formalisms have identified certain key principles \citep{Joshi90}.  Firstly, that the set of rules in a given grammar is bounded.  This in turn implies that the set of properties and relations
in a given grammar is also bounded.

But language is unbounded\footnote{A set of things (e.g.\ the sentences of a language) have unbounded size if for any finite size there is always some element in the set which is larger than that.} in nature, since sentences and texts can be arbitrarily long.  Grammar formalisms capture this unboundedness by allowing an unbounded number of entities in a representation, and thus an unbounded number of rule applications.  It is generally accepted that the number of entities grows linearly with the length of the sentence \citep{Joshi90}, so each level can have at most a number of entities which is linear in the number of entities at the level(s) below.

Computational linguistic grammar formalisms also typically assume that the properties and relations are discrete, called symbolic representations.  These may be atomic categories, as in CFGs, TAGs, CCG and dependency grammar, or they may be feature structures, as in HPSG.

\subsection{Connectionism}

Other researchers who were more interested in the computational properties of neurological systems found 
this reliance on discrete categorical representations untenable.  Processing in the brain used real-valued representations distributed across many neurons.  Based on successes following the development of multi-layered perceptrons (MLPs) \citep{Rumelhart86}, an approach to modelling cognitive phenomena was developed called connectionism.  Connectionism uses vector-space representations to reflect the distributed continuous nature of representations in the brain.  Similarly, their rules are specified with vectors of continuous parameters.  MLPs are so powerful that they are arbitrary function approximators \citep{Hornik89}.  And thanks to backpropagation learning \citep{Rumelhart86b} in neural network models, such as MLPs and Simple Recurrent Networks (SRNs) \citep{Elman90}, these vector-space representations and rules could be learned from data.  

The ability to learn powerful vector-space representations from data led many connectionist to argue that the complex discrete structured representations of computational linguistics were neither necessary nor desirable (e.g.\ \citet{Smolensky88,Smolensky90,Elman91,Miikkulainen93,Seidenberg07}).  Distributed vector-space representations were thought to be so powerful that there was no need for anything else.  Learning from data made linguistic theories irrelevant.  (See also \citep{Collobert08,Collobert11,Sutskever14} for more recent incarnations.)

The idea that vector-space representations are adequate for natural language and other cognitive phenomena was questioned from several directions.  From neuroscience, researchers questioned how a simple vector could encode features of more than one thing at a time.  If we see a red square together with a blue triangle, how do we represent the difference between that and a red triangle with a blue square, since the vector elements for red, blue, square and triangle would all be active at the same time?  This is known as the variable binding problem, so called because variables are used to do this binding in symbolic representations, as in $red(x)\wedge triangle(x)\wedge blue(y)\wedge square(y)$.
One proposal has been that the precise timing of neuron activation spikes could be used to encode variable binding, called Temporal Synchrony Variable Binding \citep{Malsburg81,Shastri93}.  Neural spike trains have both a phase and a period, so the phase could be used to encode variable binding while still allowing the period to be used for sequential computation.  This work indicated how entities could be represented in a neurally-inspired computational architecture.

The adequacy of vector-space representations was also questioned based on the regularities found in natural language.  In particular, \citet{Fodor88} argued that connectionist architectures were not adequate to account for regularities which they characterised as {\em systematicity} (see also \citep{Smolensky90,Fodor90}).  In essence, systematicity requires that learned rules generalise in a way that respects structured representations.  Here again the issue is representing multiple entities at the same time, but with the additional requirement of representing the structural relationships between these entities.  Only rules which are parameterised in terms of such representations can generalise in a way which accounts for the generalisations found in language.

Early work on neural networks for natural language recognised the significance of variable binding for solving the issues with systematicity \citep{Henderson96,Henderson00a}.  \citet{Henderson94d,Henderson00a} argued that extending neural networks with temporal synchrony variable binding made them powerful enough to account for the regularities found in language.  Using time to encode variable bindings means that learning could generalise in a linguistically appropriate way \citep{Henderson96}, since rules (neuronal synapses) learned for one variable (time) would systematically generalise to other variables.
Although relations were not stored explicitly, it was claimed that  for language understanding it is adequate to recover them from the features of the entities \citep{Henderson94d,Henderson00a}.
But these arguments were largely theoretical, and it was not clear how they could be incorporated in learning-based architectures.

\subsection{Statistical Models}

Although researchers in computational linguistics did not want to abandon their representations, they did recognise the importance of learning from data.  The first successes in this direction came from learning rules with statistical methods, such as
part-of-speech tagging with hidden Markov models. 
For syntactic parsing, the development of the Penn Treebank led to many statistical models which learned the rules of grammar \citep{Collins97,Collins99,Charniak97a,Ratnaparkhi99}.

These statistical models were very successful at learning from the distributions of linguistic representations which had been annotated in the corpus they were trained on.  But they still required linguistically-motivated designs to work well.  In particular, feature engineering is necessary to make sure that these statistical machine-learning method can search a space of rules which is sufficiently broad to include good models but sufficiently narrow to allow learning from limited data.

\section{Inducing Features of Entities}

Early work on neural networks for natural language recognised the potential of neural networks for learning the features as well, replacing feature engineering.
But empirically successful neural network models for NLP were only achieved with approaches where the neural network was used to model one component within an otherwise traditional symbolic NLP model.

The first work to achieve empirical success in comparison to non-neural statistical models was work on language modelling.  \citet{Bengio01,Bengio03} used an MLP to estimate the parameters of an n-gram language model, and showed improvements when interpolated with a statistical n-gram language model.  A crucial innovation of this model was the introduction of word embeddings.
The idea that the properties of a word could be represented by a vector reflecting the distribution of the word in text was introduced earlier in non-neural statistical models
(e.g.\ \citep{Deerwester1990,Schutze93,burgess1998simple,pado-lapata-2007-dependency,erk-2010-word}).
This work showed that similarity in the resulting vector space is correlated with semantic similarity.
Learning vector-space representations of words with neural networks (rather than SVD) have showed similar effects
(e.g.\ \citep{turian-etal-2010-word,word2vec2_nips,levy-etal-2015-improving,pennington-etal-2014-glove}),
resulting in impressive improvements for many NLP tasks.

More recent work has used neural network language models to learn context-dependent embeddings of words.
We will refer to such context-dependent embeddings as {\em token embeddings}.   For example, \citet{peters-etal-2018-deep} train a stacked BiLSTM language model,
and these token embeddings have proved effective in many tasks.
More such models will be discussed below.

\begin{table}
\centering
\begin{adjustbox}{width=\linewidth}
\begin{tabular}{|lr@{~}|@{~}l@{~~}l@{~}|@{~}l@{~}|}
  \hline
  \multicolumn{5}{|c|}{PTB Constituents}\\
  \hline
  model & & LP & LR & F1 \\
  \hline
  \citet{Costa01} & PoS & 57.8 & 64.9 & 61.1 \\
  \citet{Henderson03_naacl} & PoS  & 83.3 & 84.3 & 83.8 \\
  \hline
  \citet{Henderson03_naacl} &  & 88.8 & 89.5 & 89.1 \\ 
  \citet{Henderson04_acl} &  & 89.8 & 90.4 & 90.1 \\ 
  \citet{Vinyals15_nips} seq2seq &  &  &  & $<$70 \\
  \citet{Vinyals15_nips} attn &  &  &  & 88.3 \\
  \citet{Vinyals15_nips} seq2seq & semisup &  &  & 90.5 \\
  \hline
\end{tabular}
\end{adjustbox}
\newline
\begin{adjustbox}{width=\linewidth}
\begin{tabular}{|l@{~~}r@{~}|@{~}l@{~}|@{~}l@{~}|}
  \hline
  \multicolumn{4}{|c|}{CoNLL09 Dependencies}\\
  \hline
  model (transition-based) & & UAS & LAS \\ 
  \hline
  \citet{titov-henderson-2007-latent}* &  & 91.44 & 88.65 \\
  \citet{chen-manning-2014-fast}*  &  & 89.17 & 86.49 \\
  \citet{yazdani-henderson-2015-incremental} &  & 90.75 & 88.14 \\
  \hline\hline
  \multicolumn{4}{|c|}{Stanford Dependencies}\\
  \hline
  model (transition-based) & & UAS & LAS \\ 
  \hline
  \citet{chen-manning-2014-fast} & & 91.80 & 89.60 \\
  \citet{dyer-etal-2015-transition} & & 93.10 & 90.90 \\
  \citet{andor-etal-2016-globally} & & 94.61 & 92.79 \\
  \citet{kiperwasser-goldberg-2016-simple} &  & 93.9 & 91.9 \\
  \citet{Mohammadshahi19_arxiv} & BERT & 95.63 & 93.81 \\
  \hline
\end{tabular}
\end{adjustbox}
  \vspace{-1ex}
  \caption{\label{tab:transition_parsing} Some neural network parsing results on Penn Treebank WSJ. LP/LR/F1: labelled constituent precision/recall/F-measure. UAS/LAS: unlabelled/labelled dependency accuracy.
    ~*\,results reported in \citep{yazdani-henderson-2015-incremental}.
  \vspace{-1ex}
}
\end{table}

For syntactic parsing, early connectionist approaches \citep{Jain91,Miikkulainen93,Ho99,Costa01} had limited success.
The first neural network models to achieve empirical success used a recurrent neural network to model the derivation structure of a traditional syntactic constituency parser \citep{Henderson03_naacl,Henderson04_acl}.  The recurrent neural network learns to model the sequence of parser actions, estimating the probability of the next parser action given the history of previous parser actions.  This allows the decoding algorithm from the traditional parsing model to be used to efficiently search the space of possible parses.
These models have also been applied to syntactic dependency parsing \citep{Titov07_iwpt,yazdani-henderson-2015-incremental} and joint syntactic-semantic dependency parsing \citep{Henderson13_cl}.

Crucially, these neural networks do not model the sequence of parser decisions as a flat sequence, but instead model the derivation structure it specifies.  A derivation structure includes relationships for the inter-dependencies between nodes in the parse tree.  The pattern of interconnections between hidden layers of the recurrent neural network (henceforth referred to as the {\em model structure}) is designed to follow locality in this derivation structure, thereby giving the neural network a linguistically appropriate inductive bias.
More recently, \citet{dyer-etal-2015-transition} provide a more direct
relationship between the derivation structure and the model structure with their StackLSTM parsing model. 

In all these models, the use of recurrent neural networks allows arbitrarily large parse structures to be modelled without making any hard independence assumptions, in contrast to non-neural statistical models.
Feed-forward neural networks have also been applied to modelling the derivation structure \citep{chen-manning-2014-fast}, but the accuracy is worse than using recurrent models (see Table~\ref{tab:transition_parsing}), presumably because such models suffer from the need to make hard independence assumptions.

Representing the parse tree as a derivation sequence, rather than a derivation structure, makes it possible to define syntactic parsing as a sequence-to-sequence problem, mapping the sentence to its parse sequence.
If a neural network architecture for modelling sequences (called {\em seq2seq} models) can perform well at this task, then maybe the structured linguistic representations of natural language are not necessary (contrary to \citet{Fodor88}), not even to predict those structures.
\citet{Vinyals15_nips} report very poor results for seq2seq models when trained on the standard dataset, but good results when trained on very large automatically-parsed corpora (see Table~\ref{tab:transition_parsing} {\em semisup}).
They only achieve good results with the limited standard dataset by adding attention, which we will argue below makes the model no longer a seq2seq model.
This indicates that structured representations really do capture important generalisations about language.\footnote{See \citep{Collobert08,Collobert11} for an earlier related line of work.}

In contrast to seq2seq models, there have also been neural network models of parsing which directly represent linguistic structure, rather than just derivation structure, giving them induced vector representations which map one-to-one with the entities in the linguistic representation.  Typically, a recursive neural network is used to compute embeddings of syntactic constituents bottom-up. 
\citet{dyer-etal-2015-transition} showed improvements by adding these representations to a model of the derivation structure.  
\citet{socher-etal-2013-parsing} only modelled the linguistic structure, making it difficult to do decoding efficiently.
But the resulting induced constituent embeddings have a clear linguistic interpretation, making it easier to use them within other tasks, such as sentiment analysis \citep{Socher13_emnlp}.
Similarly, models based on Graph Convolutional Networks have induced embeddings with clear linguistic interpretations within pre-defined model structures (e.g.\ \citep{marcheggiani-titov-2017-encoding,marcheggiani-etal-2018-exploiting}).

All these results demonstrate the incredible effectiveness of inducing vector-space representations with neural networks,
relieving us from the need to do feature engineering.  But neural networks do not relieve us of the need to understand the nature of language when designing our models.  Instead of feature engineering, these results show that the best accuracy is achieved by engineering the inductive bias of deep learning models through their model structure.  By designing a hand-coded model structure which reflects the linguistic structure, locality in the model structure can reflect locality in the linguistic structure.  The neural network then induces features of the entities in this model structure.

\section{Inducing Relations between Entities}

With the introduction of attention-based models, the model structure can now be learned.  By choosing the nodes to be linguistically-motivated entities, learning the model structure in effect learns the statistical inter-dependencies between entities, which is what we have been referring to as relations.

\subsection{Attention-Based Models and Variable Binding}

The first proposal of an attention-based neural model learned a soft alignment between the target and source words in neural machine translation (NMT) \citep{Bahdanau15}.
The model structure of the source sentence encoder and the model structure of the target sentence decoder are both flat sequences, but when each target word is generated, it computes attention weights over all source words.  These attention weights directly express how target words are correlated with source words, and in this sense can be seen as a soft version of the alignment structure.  In traditional statistical machine translation, this alignment structure is determined with a separate alignment algorithm, and then frozen while training the model.  In contrast, the attention-based NMT model learns the alignment structure jointly with learning the encoder and decoder, inside the deep learning architecture \citep{Bahdanau15}.

This attention-based approach to NMT was also applied to mapping a sentence to its syntactic parse \citep{Vinyals15_nips}.  The attention function learns the structure of the relationship between the sentence and its syntactic derivation sequence, but does not have any representation of the structure of the syntactic derivation itself.  Empirical results are much better than their seq2seq model \citep{Vinyals15_nips}, but not as good as models which explicitly model both structures (see Table~\ref{tab:transition_parsing}).

The change from the sequential LSTM decoders of previous NMT models to LSTM decoders with attention seems like a simple addition, but it fundamentally changes the kinds of generalisations which the model is able to learn.  At each step in decoding, the state of a sequential LSTM model is a single vector, whereas adding attention means that the state needs to include the unboundedly large set of vectors being attended to.  This use of an unbounded state is more similar to the above models with predefined model structure, where an unboundedly large stack is needed to specify the parser state.  This change in representation leads to a profound change in the generalisations which can be learned.  Parameterised rules which are learned when paying attention to one of these vectors (in the set or in the stack) automatically generalise to the other vectors.  In other words, attention-based models have variable binding,
which sequential LSTMs do not.  Each vector represents the features for one entity, multiple entities can be kept in memory at the same time, and rules generalise across these entities.  In this sense it is wrong to refer to attention-based models as sequence models; they are in fact {\em induced-structure} models.  We will expand on this perspective in the rest of this section.

\subsection{Transformer and Systematicity}

The generality of attention as a structure-induction method soon became apparent,
culminated in the development of the Transformer architecture \cite{Vaswani_NIPS2017}.  
Transformer has multiple stacked layers of self-attention (attention to the other words in the same sequence), interleaved with nonlinear functions applied to individual vectors.  Each attention layer has multiple attention heads, allowing each head to learn a different type of relation.  A Transformer-encoder has one column of stacked vectors for each position in the input sequence, and the model parameters are shared across positions.  A Transformer-decoder adds attention over an encoded text, and predicts words one at a time after encoding the prefix of previously generated words.

Although it was developed for encoding and generating sequences, in Transformer the sequential structure is not hard-coded into the model structure, unlike previous models of deep learning for sequences (e.g.\ LSTMs \citep{LSTM97} and CNNs \citep{LeCun95}).  Instead, the sequential structure is input in the form of position embeddings.  In our formulation, position embeddings are just properties of individual entities (typically words or subwords).  As such, these inputs facilitate learning about absolute positions.  But they are also designed to allow the model to easily calculate relative position between entities.  This allows the model's attention functions to learn to discover the relative position structure of the underlying sequence.  In fact, explicitly  inputting relative position relations as embeddings into the attention functions works even better \citep{shaw-etal-2018-self} (discussed further below).   Whether input as properties or as relations, these inputs are just features, not hard-coded model structure.  The attention weight functions can then learn to use these features to induce their own structure.

The appropriateness and generality for natural language of the Transformer architecture became even more apparent with the development of pretrained Transformer models like BERT \citep{devlin-etal-2019-bert}.
BERT models are large Transformer models trained mostly on a masked language model objective, as well as a next-sentence prediction objective.
After training on a very large amount of unlabelled text, the resulting pretrained model can be fine tuned for various tasks, with very impressive improvements in accuracy across a wide variety of tasks.
The success of BERT has led to various analyses of what it has learned, including the structural relations learned by the attention functions.  Although there is no exact mapping from these structures to the structures posited by linguistics, there are clear indications that the attention functions are learning to extract linguistic relations \citep{voita-etal-2019-analyzing,tenney-etal-2019-bert,Reif_NIPS2019}.

With variable binding for the properties of entities and attention functions for relations between entities, Transformer can represent the kinds of structured representations argued for above.  With parameters shared across entities and sensitive to these properties and relations, learned rules are parameterised in terms of these structures.  Thus Transformer is a deep learning architecture with the kind of generalisation ability required to exhibit systematicity, as in \citep{Fodor88}.

Interestingly, the relations are not stored explicitly.  Instead they are extracted from pairs of vectors by the attention functions, as with the use of position embeddings to compute relative position relations.  For the model to induce its own structure, lower levels must learn to embed its relations in pairs of token embeddings, which higher levels of attention then extract.

That Transformer learns to embed relations in pairs of token embeddings is apparent from recent work on dependency parsing \citep{kondratyuk-straka-2019-75,Mohammadshahi19_arxiv,Mohammadshahi20_arxiv}.  Earlier models of dependency parsing successfully use BiLSTMs to embed syntactic dependencies in pairs of token embeddings (e.g.\ \citep{kiperwasser-goldberg-2016-simple,Dozat16}), which are then extracted
to predict the dependency tree.  \citet{Mohammadshahi19_arxiv,Mohammadshahi20_arxiv} use their proposed Graph-to-Graph Transformer to encode dependencies in pairs of token embeddings, for transition-based and graph-based dependency parsing respectively.  Graph-to-Graph Transformer also inputs previously predicted dependency relations into its attention functions (like relative position encoding \citep{shaw-etal-2018-self}).   These parsers achieve state of the art accuracies,
indicating that Transformer finds it easy to input and predict syntactic dependency relations via pairs of token embeddings.  Interestingly, initialising the model with pretrained BERT results in large improvements, indicating that BERT representations also encode syntactically-relevant relations in pairs of token embeddings.

\subsection{Nonparametric Representations}

As we have seen, the problem with vector-space models is not simply about representations, but about the way learned rules generalise.  In work on grammar formalisms, generalisation is analysed by looking at the unbounded case, since any bounded case can simply be memorised.  But the use of continuous representations
does not fit well with the theory of grammar formalisms, which assumes a bounded vocabulary of atomic categories.  Instead we propose an analysis of the generalisation abilities of Transformer in terms of  theory from machine learning, Bayesian nonparametric learning \cite{Jordan10}.  We argue that the representations of Transformer are the minimal nonparametric extension of a vector space.

To connect Transformer to Bayesian probabilities, we assume that a Transformer representation can be thought of as the parameters of a probability distribution.  This is natural, since a model's state represents a belief about the input, and in Bayesian approaches beliefs are probability distributions.  From this perspective, computing a representation is inferring the parameters of a probability distribution from the observed input.  This is analogous to Bayesian learning,
where we infer the parameters of a distribution over models from observed training data.
In this section, we outline how theory from Bayesian learning helps us understand
how the representations of Transformer lead to better generalisation.

We do not make any specific assumptions about what probability distributions are specified by a Transformer representation, but it is useful to keep in mind an example.  One possibility is a mixture model, where each vector specifies the parameters of a multi-dimensional distribution, and the total distribution is the weighted sum across the vectors of these distributions.  For example, we can interpret the vectors $x{=}x_1,\ldots ,x_n$ in a Transformer's representation as specifying a belief about the queries $q$ that will be received from a downstream attention function, as in:
\vspace{-1ex}
\begin{align*}
  P(q|x) =~& \sum_i P(i|x)\, P(q|x_i) \\[-0.7ex]
  P(i|x) =~& \exp( {\tfrac{1}{2}} ||x_i||^2 ) ~/~ \sum_i \exp( \tfrac{1}{2} ||x_i||^2 ) \\[-0.3ex] 
  P(q|x_i) =~& \mathcal{N}(q\, ;\, \mu{=}x_i,\sigma{=}1)
\\[-4ex]
\end{align*}
With this interpretation of $x$, we can use the fact that
$P(i|x,q) \propto P(i|x)\, P(q|x_i) \propto \exp( q\!\cdot\! x_i )$
(ignoring factors independent of $i$) to reinterpret a standard attention function.
\ignore{
\begin{align*}
  P(i|x,q) =~& P(i|x) P(q|x_i) / (\sum_i P(i|x) P(q|x_i)) \\
=~& P(i|x) \exp(-(x_i-q)^2/2) / 2.507
  / ( \sum_i P(i|x) \exp(-(x_i-q)^2/2) / 2.507 ) \\
=~& P(i|x) \exp( -(||x_i||^2 + -2x_i\cdot q + ||q||^2) / 2 )
  / ( \sum_i P(i|x) \exp( -(||x_i||^2 + -2x_i\cdot q + ||q||^2) / 2 ) ) \\
=~& P(i|x) \exp( (-||x_i||^2/2 + x_i\cdot q) )
  / ( \sum_i P(i|x) \exp( (-||x_i||^2/2 + x_i\cdot q) ) ) \\
=~& P(i|x) ( \exp( -||x_i||^2 / 2 ) ) \exp( x_i\cdot q ) 
  / ( \sum_i P(i|x) ( \exp( -||x_i||^2 / 2 ) ) \exp( x_i\cdot q )  ) \\
=~& \exp( x_i\cdot q ) 
  / ( \sum_i \exp( x_i\cdot q ) )
  \end{align*}
}
\ignore{
\begin{align*}
  P(q|x) =~& \sum_i P(i|x)\, P(q|x_i) \\
  P(i|x) \propto~& ||x_i|| \\
  P(q|x_i) \propto~& \cos(q,\, x_i)
\\[-4ex]
\end{align*}
With this interpretation of $x$, we can use the fact that
$P(i|x,q) \propto P(i|x)\, P(q|x_i) \propto q\!\cdot\! x_i$
(ignoring factors independent of $i$) to reinterpret a simple attention function.
}

Since Transformer has a discrete segmentation of its representation into positions (which we call entities), but no explicit representation of structure, we can think of this representation as a bag of vectors (BoV, i.e.\ a set of instances of vectors).
Each layer has a BoV representation, which is aligned with the BoV representation below it.  The final output only becomes a sequence if the downstream task imposes explicit sequential structure on it, which attention alone does not.

These bag of vector representations have two very interesting properties for natural language.  First, the number of vectors in the bag can grow arbitrarily large, which captures the unbounded nature of language.
Secondly, the vectors in the bag are {\em exchangeable}, in the sense of \citet{Jordan10}.
In other words, renumbering the indices used to refer to the different vectors will not change the interpretation of the representation.\footnote{These indices should not be confused with position embeddings.  In fact, position embeddings are needed precisely because the indices are meaningless to the model.}  This is because the learned parameters in Transformer are shared across all positions.  These two properties are clearly related; exchangeability allows learning to generalise to unbounded representations, since there is no need to learn about indices which are not in the training data.

These properties mean that BoV representations are nonparametric representations.
In other words, the specification of a BoV representation cannot be done just by choosing values for a fixed set of parameters.  The number of parameters you need grows with the size of the bag.
This is crucial for language because the amount of information conveyed by a text grows with the length of the text, so we need nonparametric representations.

To illustrate the usefulness of this view of BoVs as nonparametric representations, we propose to use methods from Bayesian learning to define a prior distribution over BoVs where the size of the bag is not known.  Such a prior would be needed for learning the number of entities in a Transformer representation, discussed below, using variational Bayesian approaches.  For this example, we will use the above interpretation of a BoV $x{=}\{x_i\,|\,1{\leq}i{\leq}k\}$ as specifying a distribution over queries, $P(q|x){=}\sum_i P(i|x)P(q|x_i)$.  A prior distribution over these $P(q|x)$ distributions can be specified, for example, with a Dirichlet Process, $DP(\alpha,G_0)$.  The concentration parameter $\alpha$ controls the generation of a sequence of probabilities $\rho_1,\rho_2,\ldots$, which correspond to the $P(i|x)$ distribution (parameterised by the
$||x_i||$).  The base distribution $G_0$ controls the generation of the $P(q|x_i)$ distributions (parameterised by the 
$x_i$).  

The use of exchangeability to support generalisation to unbounded representations implies a third interesting property, discrete segmentation into entities.
In other words, the information in a BoV is spread across an integer number of vectors.
A vector cannot be half included in a BoV; it is either included or not.
In changing from a vector space to a bag-of-vector space,
the only change is this discrete segmentation into entities.
In particular, no discrete representation of structure is added to the representation.
Thus, the BoV representation of Transformer is the minimal nonparametric extension of a vector space.

With this minimal nonparametric extension, Transformer is able to explicitly represent entities and their properties, and implicitly represent a structure of relations between these entities.  The continuing astounding success of Transformer in natural language understanding tasks suggests that this is an adequate deep learning architecture for the kinds of structured representations needed to account for the nature of language.

\section{Looking Forward: Inducing Levels and their Entities}

As argued above, the great success of neural networks in NLP has not been because they are radically different from pre-neural computational theories of language, but because they have succeeded in replacing hand-coded components of those models with learned components which are specifically designed to capture the same generalisations.  We predict that there is at least one more hand-coded aspect of these models which can be learned from data, but question whether they all can be.

Transformer can learn representations of entities and their relations, but 
current work (to the best of our knowledge) all assumes that the set of entities is a predefined function of the text.
Given a sentence, a Transformer does not learn how many vectors it should use to represent it.  The number of positions in the input sequence is given, and the number of token embeddings is the same as the number of input positions.  When a Transformer decoder generates a sentence, the number of positions is chosen by the model, but it is simply trying to guess the number of positions that would have been given if this was a training example.  These Transformer models never try to induce the number of token embeddings they use in an unsupervised way.\footnote{Recent work on inducing sparsity in attention weights \citep{correia-etal-2019-adaptively} effectively learns to reduce the number of entities used by individual attention heads, but not by the model as a whole.}

Given that current models hard-code different token definitions for different tasks (e.g. character embeddings versus word embeddings versus sentence embeddings), it is natural to ask whether a specification of the set of entities at a given level of representation can be learned.
There are models which induce the set of entities in an input text, but these are (to the best of our knowledge) not learned jointly with a downstream deep learning model.
Common examples include BPE \citep{sennrich-etal-2016-neural} and unigram language model \citep{kudo-2018-subword}, which use statistics of character n-grams to decide how to split words into subwords.  The resulting subwords then become the entities for a deep learning model, such as Transformer (e.g.\ BERT), but they do not explicitly optimise the performance of this downstream model.
In a more linguistically-informed approach to the same problem, statistical models have been proposed for morphology induction (e.g.\ \citep{Elsner13}).
Also, Semi-Markov CRF models \citep{Sarawagi05} can learn segmentations of an input string, which have been used in the output layers of neural models (e.g.\ \citep{kong2015segmental}).
The success of these models in finding useful segmentations of characters into subwords suggests that learning the set  of entities can be integrated into a deep learning model.  But this task is complicated by the inherently discrete nature of the segmentation into entities.  It remains to find effective neural architectures for learning the set of entities jointly with the rest of the neural model, and for generalising such methods from the level of character strings to higher levels of representation.

The other remaining hand-coded component of computational linguistic models is levels of representation.  Neural network models of language typically only represent a few levels, such as the character sequence plus the word sequence, the word sequence plus the syntax tree, or the word sequence plus the syntax tree plus the predicate-argument structure \cite{Henderson13_cl,swayamdipta-etal-2016-greedy}.  And these levels and their entities are defined before training starts, either in pre-processing or in annotated data.  If we had methods for inducing the set of entities at a given level (discussed above), then we could begin to ask whether we can induce the levels themselves.

One common approach to inducing levels of representation in neural models is to deny it is a problem.  Seq2seq and end2end models typically take this approach.  These models only include representations at a lower level, both for input and output, and try to achieve equivalent performance to models which postulate some higher level of representation (e.g.\ \citep{Collobert08,Collobert11,Sutskever14,Vinyals15_nips}). 
The most successful example of this approach has been neural machine translation.  The ability of neural networks to learn such models is impressive, but the challenge of general natural language understanding is much greater than machine translation.
Nonetheless, models which do not explicitly model levels of representation can show that they have learned about different levels implicitly \citep{peters-etal-2018-deep,tenney-etal-2019-bert}.

We think that it is far more likely that we will be able to design neural architectures which induce multiple levels of representation than it is that we can ignore this problem entirely.  However, it is not at all clear that even this will be possible.
Unlike the components previously learned, no linguistic theory postulates different levels of representation for different languages.
Generally speaking, there is a consensus that the levels minimally include phonology, morphology, syntactic structure, predicate-argument structure, and discourse structure.
This language-universal nature of levels of representation suggests that in humans the levels of linguistic representation are innate.  This draws into question whether levels of representation can be learned at all.  Perhaps they are innate because human brains are not able to learn them from data.  If so, perhaps it is the same for neural networks, and so attempts to induce levels of representation are doomed to failure.

Or perhaps we can find new neural network architectures which are even more powerful than what is now thought possible.  It wouldn't be the first time!

\section{Conclusions}

We conclude that the nature of language has influenced the design of deep learning architectures in fundamental ways.  Vector space representations (as in MLPs) are not adequate, nor are vector spaces which evolve over time (as in LSTMs).  Attention-based models are fundamentally different because they use bag-of-vector representations.  BoV representations are nonparametric representations, in that the number of vectors in the bag can grow arbitrarily large, and these vectors are exchangeable.

With BoV representations, attention-based neural network models like Transformer can model the kinds of unbounded structured representations that computational linguists have found to be necessary to capture the generalisations in natural language.
And deep learning
allows many aspects of these structured representations to be learned from data.

However, successful deep learning architectures for natural language currently still have many hand-coded aspects.  The levels of representation are hand-coded, based on linguistic theory or available resources.  Often deep learning models only address one level at a time, whereas a full model would involve levels ranging from the perceptual input to logical reasoning.  Even within a given level, the set of entities is a pre-defined function of the text.

This analysis suggests that an important next step in deep learning architectures for natural language understanding will be the induction of entities.  It is not clear what advances in deep learning methods will be necessary to improve over our current fixed entity definitions, nor whether the resulting entities will be any different from the ones postulated by linguistic theory.  If we can induce the entities at a given level, a more challenging task will be the induction of the levels themselves.  The presumably-innate nature of linguistic levels suggests that this might not even be possible.

But of one thing we can be certain: the immense success of adapting deep learning architectures to fit with our computational-linguistic understanding of the nature of language will doubtless continue, with greater insights for both natural language processing and machine learning.

\section*{Acknowledgements}

We would like to thank Paola Merlo, Suzanne Stevenson, Ivan Titov, members of the Idiap NLU group, and the anonymous reviewers for their comments and suggestions.

\bibliography{anthology,mybib,arm_acl2020}

\begin{thebibliography}{79}
\expandafter\ifx\csname natexlab\endcsname\relax\def\natexlab#1{#1}\fi

\bibitem[{Andor et~al.(2016)Andor, Alberti, Weiss, Severyn, Presta, Ganchev,
  Petrov, and Collins}]{andor-etal-2016-globally}
Daniel Andor, Chris Alberti, David Weiss, Aliaksei Severyn, Alessandro Presta,
  Kuzman Ganchev, Slav Petrov, and Michael Collins. 2016.
\newblock \href {https://doi.org/10.18653/v1/P16-1231} {Globally normalized
  transition-based neural networks}.
\newblock In \emph{Proceedings of the 54th Annual Meeting of the Association
  for Computational Linguistics (Volume 1: Long Papers)}, pages 2442--2452,
  Berlin, Germany. Association for Computational Linguistics.

\bibitem[{Bahdanau et~al.(2015)Bahdanau, Cho, and Bengio}]{Bahdanau15}
Dzmitry Bahdanau, Kyunghyun Cho, and Yoshua Bengio. 2015.
\newblock Neural machine translation by jointly learning to align and
  translate.
\newblock In \emph{Proceedings of ICLR}.

\bibitem[{Bengio et~al.(2001)Bengio, Ducharme, and Vincent}]{Bengio01}
Yoshua Bengio, R\'ejean Ducharme, and Pascal Vincent. 2001.
\newblock A neural probabilistic language model.
\newblock In \emph{Advances in Neural Information Processing Systems 13}, pages
  932--938. MIT Press.

\bibitem[{Bengio et~al.(2003)Bengio, Ducharme, Vincent, and Janvin}]{Bengio03}
Yoshua Bengio, R\'{e}jean Ducharme, Pascal Vincent, and Christian Janvin. 2003.
\newblock A neural probabilistic language model.
\newblock \emph{J. Machine Learning Research}, 3:1137--1155.

\bibitem[{Burgess(1998)}]{burgess1998simple}
Curt Burgess. 1998.
\newblock From simple associations to the building blocks of language: Modeling
  meaning in memory with the {HAL} model.
\newblock \emph{Behavior Research Methods, Instruments, \& Computers},
  30(2):188--198.

\bibitem[{Charniak(1997)}]{Charniak97a}
Eugene Charniak. 1997.
\newblock Statistical parsing with a context-free grammar and word statistics.
\newblock In \emph{Proc. 14th National Conference on Artificial Intelligence},
  Providence, RI. AAAI Press/MIT Press.

\bibitem[{Chen and Manning(2014)}]{chen-manning-2014-fast}
Danqi Chen and Christopher Manning. 2014.
\newblock \href {https://doi.org/10.3115/v1/D14-1082} {A fast and accurate
  dependency parser using neural networks}.
\newblock In \emph{Proceedings of the 2014 Conference on Empirical Methods in
  Natural Language Processing ({EMNLP})}, pages 740--750, Doha, Qatar.
  Association for Computational Linguistics.

\bibitem[{Chomsky(1959)}]{Chomsky59}
Noam Chomsky. 1959.
\newblock On certain formal properties of grammars.
\newblock \emph{Information and Control}, 2:137--167.

\bibitem[{Collins(1997)}]{Collins97}
Michael Collins. 1997.
\newblock Three generative, lexicalized models for statistical parsing.
\newblock In \emph{Proc. 35th Meeting of {A}ssociation for {C}omputational
  {L}inguistics and 8th Conf. of {E}uropean Chapter of {A}ssociation for
  {C}omputational {L}inguistics}, pages 16--23, Somerset, New Jersey.

\bibitem[{Collins(1999)}]{Collins99}
Michael Collins. 1999.
\newblock \emph{Head-Driven Statistical Models for Natural Language Parsing}.
\newblock Ph.D. thesis, University of Pennsylvania, Philadelphia, PA.

\bibitem[{Collobert et~al.(2011)Collobert, Weston, Bottou, Karlen, Kavukcuoglu,
  and Kuksa}]{Collobert11}
R.~Collobert, J.~Weston, L.~Bottou, M.~Karlen, K.~Kavukcuoglu, and P.~Kuksa.
  2011.
\newblock Natural language processing (almost) from scratch.
\newblock \emph{Journal of Machine Learning Research}, 12:2493--2537.

\bibitem[{Collobert and Weston(2008)}]{Collobert08}
Ronan Collobert and Jason Weston. 2008.
\newblock A unified architecture for natural language processing: deep neural
  networks with multitask learning.
\newblock In \emph{Proceedings of the Twenty-Fifth International Conference
  (ICML 2008)}, pages 160--167, Helsinki, Finland.

\bibitem[{Correia et~al.(2019)Correia, Niculae, and
  Martins}]{correia-etal-2019-adaptively}
Gon{\c{c}}alo~M. Correia, Vlad Niculae, and Andr{\'e} F.~T. Martins. 2019.
\newblock \href {https://doi.org/10.18653/v1/D19-1223} {Adaptively sparse
  transformers}.
\newblock In \emph{Proceedings of the 2019 Conference on Empirical Methods in
  Natural Language Processing and the 9th International Joint Conference on
  Natural Language Processing (EMNLP-IJCNLP)}, pages 2174--2184, Hong Kong,
  China. Association for Computational Linguistics.

\bibitem[{Costa et~al.(2001)Costa, Lombardo, Frasconi, and Soda}]{Costa01}
Fabrizio Costa, Vincenzo Lombardo, Paolo Frasconi, and Giovanni Soda. 2001.
\newblock \href {https://doi.org/10.1007/3-540-45411-X_30} {Wide coverage
  incremental parsing by learning attachment preferences}.
\newblock pages 297--307.

\bibitem[{Deerwester et~al.(1990)Deerwester, Dumais, Furnas, Landauer, and
  Harshman}]{Deerwester1990}
Scott Deerwester, Susan~T. Dumais, George~W. Furnas, Thomas~K. Landauer, and
  Richard Harshman. 1990.
\newblock Indexing by latent semantic analysis.
\newblock \emph{Journal of the American Society for Information Science},
  41(6):391--407.

\bibitem[{Devlin et~al.(2019)Devlin, Chang, Lee, and
  Toutanova}]{devlin-etal-2019-bert}
Jacob Devlin, Ming-Wei Chang, Kenton Lee, and Kristina Toutanova. 2019.
\newblock \href {https://doi.org/10.18653/v1/N19-1423} {{BERT}: Pre-training of
  deep bidirectional transformers for language understanding}.
\newblock In \emph{Proceedings of the 2019 Conference of the North {A}merican
  Chapter of the Association for Computational Linguistics: Human Language
  Technologies, Volume 1 (Long and Short Papers)}, pages 4171--4186,
  Minneapolis, Minnesota. Association for Computational Linguistics.

\bibitem[{Dozat and Manning(2016)}]{Dozat16}
Timothy Dozat and Christopher~D. Manning. 2016.
\newblock \href {http://arxiv.org/abs/1611.01734} {Deep biaffine attention for
  neural dependency parsing}.
\newblock \emph{CoRR}, abs/1611.01734.
\newblock ICLR 2017.

\bibitem[{Dyer et~al.(2015)Dyer, Ballesteros, Ling, Matthews, and
  Smith}]{dyer-etal-2015-transition}
Chris Dyer, Miguel Ballesteros, Wang Ling, Austin Matthews, and Noah~A. Smith.
  2015.
\newblock \href {https://doi.org/10.3115/v1/P15-1033} {Transition-based
  dependency parsing with stack long short-term memory}.
\newblock In \emph{Proceedings of the 53rd Annual Meeting of the Association
  for Computational Linguistics and the 7th International Joint Conference on
  Natural Language Processing (Volume 1: Long Papers)}, pages 334--343,
  Beijing, China. Association for Computational Linguistics.

\bibitem[{Elman(1990)}]{Elman90}
Jeffrey~L. Elman. 1990.
\newblock Finding structure in time.
\newblock \emph{Cognitive Science}, 14(2):179--212.

\bibitem[{Elman(1991)}]{Elman91}
Jeffrey~L. Elman. 1991.
\newblock Distributed representations, simple recurrent networks, and
  grammatical structure.
\newblock \emph{Machine Learning}, 7:195--225.

\bibitem[{Elsner et~al.(2013)Elsner, Goldwater, Feldman, and Wood}]{Elsner13}
Micha Elsner, Sharon Goldwater, Naomi Feldman, and Frank Wood. 2013.
\newblock \href {https://www.aclweb.org/anthology/D13-1005} {A joint learning
  model of word segmentation, lexical acquisition, and phonetic variability}.
\newblock In \emph{Proceedings of the 2013 Conference on Empirical Methods in
  Natural Language Processing}, pages 42--54, Seattle, Washington, USA.
  Association for Computational Linguistics.

\bibitem[{Erk(2010)}]{erk-2010-word}
Katrin Erk. 2010.
\newblock \href {https://www.aclweb.org/anthology/W10-2803} {What is word
  meaning, really? (and how can distributional models help us describe it?)}.
\newblock In \emph{Proceedings of the 2010 Workshop on {GE}ometrical Models of
  Natural Language Semantics}, pages 17--26, Uppsala, Sweden. Association for
  Computational Linguistics.

\bibitem[{Fodor and McLaughlin(1990)}]{Fodor90}
Jerry~A. Fodor and B.~McLaughlin. 1990.
\newblock Connectionism and the problem of systematicity: Why smolensky's
  solution doesn't work.
\newblock \emph{Cognition}, 35:183--204.

\bibitem[{Fodor and Pylyshyn(1988)}]{Fodor88}
Jerry~A. Fodor and Zenon~W. Pylyshyn. 1988.
\newblock Connectionism and cognitive architecture: {A} critical analysis.
\newblock \emph{Cognition}, 28:3--71.

\bibitem[{Henderson(1994)}]{Henderson94d}
James Henderson. 1994.
\newblock \emph{Description Based Parsing in a Connectionist Network}.
\newblock Ph.D. thesis, University of Pennsylvania, Philadelphia, PA.
\newblock Technical Report MS-CIS-94-46.

\bibitem[{Henderson(1996)}]{Henderson96}
James Henderson. 1996.
\newblock A connectionist architecture with inherent systematicity.
\newblock In \emph{Proceedings of the Eighteenth Conference of the Cognitive
  Science Society}, pages 574--579, La Jolla, CA.

\bibitem[{Henderson(2000)}]{Henderson00a}
James Henderson. 2000.
\newblock Constituency, context, and connectionism in syntactic parsing.
\newblock In Matthew Crocker, Martin Pickering, and Charles Clifton, editors,
  \emph{Architectures and Mechanisms for Language Processing}, pages 189--209.
  Cambridge University Press, Cambridge UK.

\bibitem[{Henderson(2003)}]{Henderson03_naacl}
James Henderson. 2003.
\newblock Inducing history representations for broad coverage statistical
  parsing.
\newblock In \emph{Proc.\ joint meeting of North American Chapter of the
  Association for Computational Linguistics and the Human Language Technology
  Conf.}, pages 103--110, Edmonton, Canada.

\bibitem[{Henderson(2004)}]{Henderson04_acl}
James Henderson. 2004.
\newblock Discriminative training of a neural network statistical parser.
\newblock In \emph{Proceedings of the 42nd Meeting of the Association for
  Computational Linguistics (ACL'04), Main Volume}, pages 95--102, Barcelona,
  Spain.

\bibitem[{Henderson et~al.(2013)Henderson, Merlo, Titov, and
  Musillo}]{Henderson13_cl}
James Henderson, Paola Merlo, Ivan Titov, and Gabriele Musillo. 2013.
\newblock Multilingual joint parsing of syntactic and semantic dependencies
  with a latent variable model.
\newblock \emph{Computational Linguistics}, 39(4):949--998.

\bibitem[{Ho and Chan(1999)}]{Ho99}
E.K.S. Ho and L.W. Chan. 1999.
\newblock How to design a connectionist holistic parser.
\newblock \emph{Neural Computation}, 11(8):1995--2016.

\bibitem[{Hochreiter and Schmidhuber(1997)}]{LSTM97}
Sepp Hochreiter and Jürgen Schmidhuber. 1997.
\newblock \href {https://doi.org/10.1162/neco.1997.9.8.1735} {Long short-term
  memory}.
\newblock \emph{Neural Computation}, 9(8):1735--1780.

\bibitem[{Hornik et~al.(1989)Hornik, Stinchcombe, and White}]{Hornik89}
K.~Hornik, M.~Stinchcombe, and H.~White. 1989.
\newblock Multilayer feedforward networks are universal approximators.
\newblock \emph{Neural Networks}, 2:359--366.

\bibitem[{Jain(1991)}]{Jain91}
Ajay~N. Jain. 1991.
\newblock \emph{PARSEC: A Connectionist Learning Architecture for Parsing
  Spoken Language}.
\newblock Ph.D. thesis, Carnegie Mellon University, Pittsburgh, PA.

\bibitem[{Jordan(2010)}]{Jordan10}
M.I. Jordan. 2010.
\newblock Bayesian nonparametric learning: Expressive priors for intelligent
  systems.
\newblock In R.~Dechter, H.~Geffner, and J.~Halpern, editors, \emph{Heuristics,
  Probability and Causality: A Tribute to Judea Pearl}, chapter~10. College
  Publications.

\bibitem[{Joshi(1987)}]{Joshi87}
Aravind~K. Joshi. 1987.
\newblock An introduction to tree adjoining grammars.
\newblock In Alexis Manaster-Ramer, editor, \emph{Mathematics of Language}.
  John Benjamins, Amsterdam.

\bibitem[{Joshi et~al.(1990)Joshi, Vijay-Shanker, and Weir}]{Joshi90}
Aravind~K. Joshi, K.~Vijay-Shanker, and David Weir. 1990.
\newblock The convergence of mildly context-sensitive grammatical formalisms.
\newblock In Peter Sells, Stuart Shieber, and Tom Wasow, editors,
  \emph{Foundational Issues in Natural Language Processing}. MIT Press,
  Cambridge MA.
\newblock Forthcoming.

\bibitem[{Kiperwasser and Goldberg(2016)}]{kiperwasser-goldberg-2016-simple}
Eliyahu Kiperwasser and Yoav Goldberg. 2016.
\newblock \href {https://doi.org/10.1162/tacl_a_00101} {Simple and accurate
  dependency parsing using bidirectional {LSTM} feature representations}.
\newblock \emph{Transactions of the Association for Computational Linguistics},
  4:313--327.

\bibitem[{Kondratyuk and Straka(2019)}]{kondratyuk-straka-2019-75}
Dan Kondratyuk and Milan Straka. 2019.
\newblock \href {https://doi.org/10.18653/v1/D19-1279} {75 languages, 1 model:
  Parsing universal dependencies universally}.
\newblock In \emph{Proceedings of the 2019 Conference on Empirical Methods in
  Natural Language Processing and the 9th International Joint Conference on
  Natural Language Processing (EMNLP-IJCNLP)}, pages 2779--2795, Hong Kong,
  China. Association for Computational Linguistics.

\bibitem[{Kong et~al.(2015)Kong, Dyer, and Smith}]{kong2015segmental}
Lingpeng Kong, Chris Dyer, and Noah~A. Smith. 2015.
\newblock \href {http://arxiv.org/abs/1511.06018} {Segmental recurrent neural
  networks}.

\bibitem[{Kudo(2018)}]{kudo-2018-subword}
Taku Kudo. 2018.
\newblock \href {https://doi.org/10.18653/v1/P18-1007} {Subword regularization:
  Improving neural network translation models with multiple subword
  candidates}.
\newblock In \emph{Proceedings of the 56th Annual Meeting of the Association
  for Computational Linguistics (Volume 1: Long Papers)}, pages 66--75,
  Melbourne, Australia. Association for Computational Linguistics.

\bibitem[{LeCun and Bengio(1995)}]{LeCun95}
Yann LeCun and Yoshua Bengio. 1995.
\newblock Convolutional networks for images, speech, and time-series.
\newblock In Michael~A. Arbib, editor, \emph{The handbook of brain theory and
  neural networks (Second ed.)}, page 276–278. MIT press.

\bibitem[{Levy et~al.(2015)Levy, Goldberg, and
  Dagan}]{levy-etal-2015-improving}
Omer Levy, Yoav Goldberg, and Ido Dagan. 2015.
\newblock \href {https://doi.org/10.1162/tacl_a_00134} {Improving
  distributional similarity with lessons learned from word embeddings}.
\newblock \emph{Transactions of the Association for Computational Linguistics},
  3:211--225.

\bibitem[{von~der Malsburg(1981)}]{Malsburg81}
C.~von~der Malsburg. 1981.
\newblock The correlation theory of brain function.
\newblock Technical Report 81-2, Max-Planck-Institute for Biophysical
  Chemistry, Gottingen.

\bibitem[{Marcheggiani et~al.(2018)Marcheggiani, Bastings, and
  Titov}]{marcheggiani-etal-2018-exploiting}
Diego Marcheggiani, Joost Bastings, and Ivan Titov. 2018.
\newblock \href {https://doi.org/10.18653/v1/N18-2078} {Exploiting semantics in
  neural machine translation with graph convolutional networks}.
\newblock In \emph{Proceedings of the 2018 Conference of the North {A}merican
  Chapter of the Association for Computational Linguistics: Human Language
  Technologies, Volume 2 (Short Papers)}, pages 486--492, New Orleans,
  Louisiana. Association for Computational Linguistics.

\bibitem[{Marcheggiani and Titov(2017)}]{marcheggiani-titov-2017-encoding}
Diego Marcheggiani and Ivan Titov. 2017.
\newblock \href {https://doi.org/10.18653/v1/D17-1159} {Encoding sentences with
  graph convolutional networks for semantic role labeling}.
\newblock In \emph{Proceedings of the 2017 Conference on Empirical Methods in
  Natural Language Processing}, pages 1506--1515, Copenhagen, Denmark.
  Association for Computational Linguistics.

\bibitem[{Miikkulainen(1993)}]{Miikkulainen93}
Risto Miikkulainen. 1993.
\newblock \emph{Subsymbolic Natural Language Processing: An integrated model of
  scripts, lexicon, and memory}.
\newblock MIT Press, Cambridge, MA.

\bibitem[{Mikolov et~al.(2013)Mikolov, Sutskever, Chen, Corrado, and
  Dean}]{word2vec2_nips}
Tomas Mikolov, Ilya Sutskever, Kai Chen, Greg~S Corrado, and Jeff Dean. 2013.
\newblock \href
  {http://papers.nips.cc/paper/5021-distributed-representations-of-words-and-phrases-and-their-compositionality.pdf}
  {Distributed representations of words and phrases and their
  compositionality}.
\newblock In C.J.C. Burges, L.~Bottou, M.~Welling, Z.~Ghahramani, and K.Q.
  Weinberger, editors, \emph{Advances in Neural Information Processing Systems
  26}, pages 3111--3119. Curran Associates, Inc.

\bibitem[{Mohammadshahi and Henderson(2019)}]{Mohammadshahi19_arxiv}
Alireza Mohammadshahi and James Henderson. 2019.
\newblock \href {http://arxiv.org/abs/1911.03561} {Graph-to-graph transformer
  for transition-based dependency parsing}.

\bibitem[{Mohammadshahi and Henderson(2020)}]{Mohammadshahi20_arxiv}
Alireza Mohammadshahi and James Henderson. 2020.
\newblock \href {http://arxiv.org/abs/2003.13118} {Recursive non-autoregressive
  graph-to-graph transformer for dependency parsing with iterative refinement}.

\bibitem[{Pad{\'o} and Lapata(2007)}]{pado-lapata-2007-dependency}
Sebastian Pad{\'o} and Mirella Lapata. 2007.
\newblock \href {https://doi.org/10.1162/coli.2007.33.2.161} {Dependency-based
  construction of semantic space models}.
\newblock \emph{Computational Linguistics}, 33(2):161--199.

\bibitem[{Pennington et~al.(2014)Pennington, Socher, and
  Manning}]{pennington-etal-2014-glove}
Jeffrey Pennington, Richard Socher, and Christopher Manning. 2014.
\newblock \href {https://doi.org/10.3115/v1/D14-1162} {{G}love: Global vectors
  for word representation}.
\newblock In \emph{Proceedings of the 2014 Conference on Empirical Methods in
  Natural Language Processing ({EMNLP})}, pages 1532--1543, Doha, Qatar.
  Association for Computational Linguistics.

\bibitem[{Peters et~al.(2018)Peters, Neumann, Iyyer, Gardner, Clark, Lee, and
  Zettlemoyer}]{peters-etal-2018-deep}
Matthew Peters, Mark Neumann, Mohit Iyyer, Matt Gardner, Christopher Clark,
  Kenton Lee, and Luke Zettlemoyer. 2018.
\newblock \href {https://doi.org/10.18653/v1/N18-1202} {Deep contextualized
  word representations}.
\newblock In \emph{Proceedings of the 2018 Conference of the North {A}merican
  Chapter of the Association for Computational Linguistics: Human Language
  Technologies, Volume 1 (Long Papers)}, pages 2227--2237, New Orleans,
  Louisiana. Association for Computational Linguistics.

\bibitem[{Pollard and Sag(1987)}]{Pollard87}
Carl Pollard and Ivan~A. Sag. 1987.
\newblock \emph{Information-Based Syntax and Semantics. Vol 1: Fundamentals}.
\newblock Center for the Study of Language and Information, Stanford, CA.

\bibitem[{Ratnaparkhi(1999)}]{Ratnaparkhi99}
Adwait Ratnaparkhi. 1999.
\newblock Learning to parse natural language with maximum entropy models.
\newblock \emph{Machine Learning}, 34:151--175.

\bibitem[{Reif et~al.(2019)Reif, Yuan, Wattenberg, Viegas, Coenen, Pearce, and
  Kim}]{Reif_NIPS2019}
Emily Reif, Ann Yuan, Martin Wattenberg, Fernanda~B Viegas, Andy Coenen, Adam
  Pearce, and Been Kim. 2019.
\newblock \href
  {http://papers.nips.cc/paper/9065-visualizing-and-measuring-the-geometry-of-bert.pdf}
  {Visualizing and measuring the geometry of bert}.
\newblock In H.~Wallach, H.~Larochelle, A.~Beygelzimer, F.~d\textquotesingle
  Alch\'{e}-Buc, E.~Fox, and R.~Garnett, editors, \emph{Advances in Neural
  Information Processing Systems 32}, pages 8594--8603. Curran Associates, Inc.

\bibitem[{Rumelhart et~al.(1986{\natexlab{a}})Rumelhart, Hinton, and
  Williams}]{Rumelhart86b}
D.~E. Rumelhart, G.~E. Hinton, and R.~J. Williams. 1986{\natexlab{a}}.
\newblock Learning internal representations by error propagation.
\newblock In D.~E. Rumelhart and J.~L. McClelland, editors, \emph{Parallel
  Distributed Processing, Vol 1}, pages 318--362. MIT Press, Cambridge, MA.

\bibitem[{Rumelhart et~al.(1986{\natexlab{b}})Rumelhart, McClelland, and the
  PDP~Reseach~group}]{Rumelhart86}
D.~E. Rumelhart, J.~L. McClelland, and the PDP~Reseach~group.
  1986{\natexlab{b}}.
\newblock \emph{Parallel Distributed Processing: Explorations in the
  microstructure of cognition, Vol 1}.
\newblock MIT Press, Cambridge, MA.

\bibitem[{Sarawagi and Cohen(2005)}]{Sarawagi05}
Sunita Sarawagi and William~W Cohen. 2005.
\newblock \href
  {http://papers.nips.cc/paper/2648-semi-markov-conditional-random-fields-for-information-extraction.pdf}
  {Semi-markov conditional random fields for information extraction}.
\newblock In L.~K. Saul, Y.~Weiss, and L.~Bottou, editors, \emph{Advances in
  Neural Information Processing Systems 17}, pages 1185--1192. MIT Press.

\bibitem[{Sch\"{u}tze(1993)}]{Schutze93}
Hinrich Sch\"{u}tze. 1993.
\newblock Word space.
\newblock In \emph{Advances in Neural Information Processing Systems 5}, pages
  895--902. Morgan Kaufmann.

\bibitem[{Seidenberg(2007)}]{Seidenberg07}
Mark~S. Seidenberg. 2007.
\newblock Connectionist models of reading.
\newblock In Gareth Gaskell, editor, \emph{Oxford Handbook of
  Psycholinguistics}, chapter~14, pages 235--250. Oxford University Press.

\bibitem[{Sennrich et~al.(2016)Sennrich, Haddow, and
  Birch}]{sennrich-etal-2016-neural}
Rico Sennrich, Barry Haddow, and Alexandra Birch. 2016.
\newblock \href {https://doi.org/10.18653/v1/P16-1162} {Neural machine
  translation of rare words with subword units}.
\newblock In \emph{Proceedings of the 54th Annual Meeting of the Association
  for Computational Linguistics (Volume 1: Long Papers)}, pages 1715--1725,
  Berlin, Germany. Association for Computational Linguistics.

\bibitem[{Shastri and Ajjanagadde(1993)}]{Shastri93}
Lokendra Shastri and Venkat Ajjanagadde. 1993.
\newblock From simple associations to systematic reasoning: A connectionist
  representation of rules, variables, and dynamic bindings using temporal
  synchrony.
\newblock \emph{Behavioral and Brain Sciences}, 16:417--451.

\bibitem[{Shaw et~al.(2018)Shaw, Uszkoreit, and Vaswani}]{shaw-etal-2018-self}
Peter Shaw, Jakob Uszkoreit, and Ashish Vaswani. 2018.
\newblock \href {https://doi.org/10.18653/v1/N18-2074} {Self-attention with
  relative position representations}.
\newblock In \emph{Proceedings of the 2018 Conference of the North {A}merican
  Chapter of the Association for Computational Linguistics: Human Language
  Technologies, Volume 2 (Short Papers)}, pages 464--468, New Orleans,
  Louisiana. Association for Computational Linguistics.

\bibitem[{Smolensky(1988)}]{Smolensky88}
Paul Smolensky. 1988.
\newblock On the proper treatment of connectionism.
\newblock \emph{Behavioral and Brain Sciences}, 11:1--17.

\bibitem[{Smolensky(1990)}]{Smolensky90}
Paul Smolensky. 1990.
\newblock Tensor product variable binding and the representation of symbolic
  structures in connectionist systems.
\newblock \emph{Artificial Intelligence}, 46(1-2):159--216.

\bibitem[{Socher et~al.(2013{\natexlab{a}})Socher, Bauer, Manning, and
  Ng}]{socher-etal-2013-parsing}
Richard Socher, John Bauer, Christopher~D. Manning, and Andrew~Y. Ng.
  2013{\natexlab{a}}.
\newblock \href {https://www.aclweb.org/anthology/P13-1045} {Parsing with
  compositional vector grammars}.
\newblock In \emph{Proceedings of the 51st Annual Meeting of the Association
  for Computational Linguistics (Volume 1: Long Papers)}, pages 455--465,
  Sofia, Bulgaria. Association for Computational Linguistics.

\bibitem[{Socher et~al.(2013{\natexlab{b}})Socher, Perelygin, Wu, Chuang,
  Manning, Ng, and Potts}]{Socher13_emnlp}
Richard Socher, Alex Perelygin, Jean Wu, Jason Chuang, Christopher~D. Manning,
  Andrew Ng, and Christopher Potts. 2013{\natexlab{b}}.
\newblock \href {http://www.aclweb.org/anthology/D13-1170} {Recursive deep
  models for semantic compositionality over a sentiment treebank}.
\newblock In \emph{Proceedings of the 2013 Conference on Empirical Methods in
  Natural Language Processing}, pages 1631--1642, Seattle, Washington, USA.
  Association for Computational Linguistics.

\bibitem[{Steedman(2000)}]{Steedman00}
Mark Steedman. 2000.
\newblock \emph{The Syntactic Process}.
\newblock MIT Press, Cambridge.

\bibitem[{Sutskever et~al.(2014)Sutskever, Vinyals, and Le}]{Sutskever14}
Ilya Sutskever, Oriol Vinyals, and Quoc~V Le. 2014.
\newblock \href
  {http://papers.nips.cc/paper/5346-sequence-to-sequence-learning-with-neural-networks.pdf}
  {Sequence to sequence learning with neural networks}.
\newblock In Z.~Ghahramani, M.~Welling, C.~Cortes, N.~D. Lawrence, and K.~Q.
  Weinberger, editors, \emph{Advances in Neural Information Processing Systems
  27}, pages 3104--3112. Curran Associates, Inc.

\bibitem[{Swayamdipta et~al.(2016)Swayamdipta, Ballesteros, Dyer, and
  Smith}]{swayamdipta-etal-2016-greedy}
Swabha Swayamdipta, Miguel Ballesteros, Chris Dyer, and Noah~A. Smith. 2016.
\newblock \href {https://doi.org/10.18653/v1/K16-1019} {Greedy, joint
  syntactic-semantic parsing with stack {LSTM}s}.
\newblock In \emph{Proceedings of The 20th {SIGNLL} Conference on Computational
  Natural Language Learning}, pages 187--197, Berlin, Germany. Association for
  Computational Linguistics.

\bibitem[{Tenney et~al.(2019)Tenney, Das, and Pavlick}]{tenney-etal-2019-bert}
Ian Tenney, Dipanjan Das, and Ellie Pavlick. 2019.
\newblock \href {https://doi.org/10.18653/v1/P19-1452} {{BERT} rediscovers the
  classical {NLP} pipeline}.
\newblock In \emph{Proceedings of the 57th Annual Meeting of the Association
  for Computational Linguistics}, pages 4593--4601, Florence, Italy.
  Association for Computational Linguistics.

\bibitem[{Titov and
  Henderson(2007{\natexlab{a}})}]{titov-henderson-2007-latent}
Ivan Titov and James Henderson. 2007{\natexlab{a}}.
\newblock \href {https://www.aclweb.org/anthology/W07-2218} {A latent variable
  model for generative dependency parsing}.
\newblock In \emph{Proceedings of the Tenth International Conference on Parsing
  Technologies}, pages 144--155, Prague, Czech Republic. Association for
  Computational Linguistics.

\bibitem[{Titov and Henderson(2007{\natexlab{b}})}]{Titov07_iwpt}
Ivan Titov and James Henderson. 2007{\natexlab{b}}.
\newblock A latent variable model for generative dependency parsing.
\newblock In \emph{Proceedings of the International Conference on Parsing
  Technologies (IWPT'07)}, Prague, Czech Republic. Association for
  Computational Linguistics.

\bibitem[{Turian et~al.(2010)Turian, Ratinov, and
  Bengio}]{turian-etal-2010-word}
Joseph Turian, Lev-Arie Ratinov, and Yoshua Bengio. 2010.
\newblock \href {https://www.aclweb.org/anthology/P10-1040} {Word
  representations: A simple and general method for semi-supervised learning}.
\newblock In \emph{Proceedings of the 48th Annual Meeting of the Association
  for Computational Linguistics}, pages 384--394, Uppsala, Sweden. Association
  for Computational Linguistics.

\bibitem[{Vaswani et~al.(2017)Vaswani, Shazeer, Parmar, Uszkoreit, Jones,
  Gomez, Kaiser, and Polosukhin}]{Vaswani_NIPS2017}
Ashish Vaswani, Noam Shazeer, Niki Parmar, Jakob Uszkoreit, Llion Jones,
  Aidan~N Gomez, \L~ukasz Kaiser, and Illia Polosukhin. 2017.
\newblock \href
  {http://papers.nips.cc/paper/7181-attention-is-all-you-need.pdf} {Attention
  is all you need}.
\newblock In I.~Guyon, U.~V. Luxburg, S.~Bengio, H.~Wallach, R.~Fergus,
  S.~Vishwanathan, and R.~Garnett, editors, \emph{Advances in Neural
  Information Processing Systems 30}, pages 5998--6008. Curran Associates, Inc.

\bibitem[{Vinyals et~al.(2015)Vinyals, Kaiser, Koo, Petrov, Sutskever, and
  Hinton}]{Vinyals15_nips}
Oriol Vinyals, \L~ukasz Kaiser, Terry Koo, Slav Petrov, Ilya Sutskever, and
  Geoffrey Hinton. 2015.
\newblock \href
  {http://papers.nips.cc/paper/5635-grammar-as-a-foreign-language.pdf} {Grammar
  as a foreign language}.
\newblock In C.~Cortes, N.~D. Lawrence, D.~D. Lee, M.~Sugiyama, and R.~Garnett,
  editors, \emph{Advances in Neural Information Processing Systems 28}, pages
  2773--2781. Curran Associates, Inc.

\bibitem[{Voita et~al.(2019)Voita, Talbot, Moiseev, Sennrich, and
  Titov}]{voita-etal-2019-analyzing}
Elena Voita, David Talbot, Fedor Moiseev, Rico Sennrich, and Ivan Titov. 2019.
\newblock \href {https://doi.org/10.18653/v1/P19-1580} {Analyzing multi-head
  self-attention: Specialized heads do the heavy lifting, the rest can be
  pruned}.
\newblock In \emph{Proceedings of the 57th Annual Meeting of the Association
  for Computational Linguistics}, pages 5797--5808, Florence, Italy.
  Association for Computational Linguistics.

\bibitem[{Yazdani and Henderson(2015)}]{yazdani-henderson-2015-incremental}
Majid Yazdani and James Henderson. 2015.
\newblock \href {https://doi.org/10.18653/v1/K15-1015} {Incremental recurrent
  neural network dependency parser with search-based discriminative training}.
\newblock In \emph{Proceedings of the Nineteenth Conference on Computational
  Natural Language Learning}, pages 142--152, Beijing, China. Association for
  Computational Linguistics.

\end{thebibliography}
\bibliographystyle{acl_natbib}

\end{document}